\documentclass[sigconf]{acmart}

\usepackage{multirow}
\usepackage{balance}
\AtBeginDocument{%
  \providecommand\BibTeX{{%
    \normalfont B\kern-0.5em{\scshape i\kern-0.25em b}\kern-0.8em\TeX}}}

\copyrightyear{2022}
\acmYear{2022}
\setcopyright{acmcopyright}
\acmConference[MM '22]{Proceedings of the 30th ACM International Conference on Multimedia}{October 10--14, 2022}{Lisboa, Portugal}
\acmBooktitle{Proceedings of the 30th ACM International Conference on Multimedia (MM '22), October 10--14, 2022, Lisboa, Portugal}
\acmPrice{15.00}
\acmDOI{10.1145/3503161.3548057}
\acmISBN{978-1-4503-9203-7/22/10}

\begin{document}

\title{CAIBC: Capturing All-round Information Beyond Color for Text-based Person Retrieval}

\author{Zijie Wang}
\affiliation{%
	\institution{Nanjing Tech University}
	\city{Nanjing}
	\country{China}}
\email{zijiewang9928@gmail.com}
\orcid{0000-0001-8739-7220}

\author{Aichun Zhu}
\authornote{Corresponding author.\vspace{-0.1cm}}
\affiliation{%
	\institution{Nanjing Tech University}
	\city{Nanjing}
	\country{China}}
\email{aichun.zhu@njtech.edu.cn}
\orcid{0000-0001-6972-5534}

\author{Jingyi Xue}
\affiliation{%
	\institution{Nanjing Tech University}
	\city{Nanjing}
	\country{China}}
\email{jyx981218@163.com}
\orcid{0000-0003-4889-6347}

\author{Xili Wan}
\affiliation{%
	\institution{Nanjing Tech University}
	\city{Nanjing}
	\country{China}}
\email{xiliwan@njtech.edu.cn}
\orcid{0000-0001-9160-8246}

\author{Chao Liu}
\affiliation{%
	\institution{Jinling Institute of Technology}
	\city{Nanjing}
	\country{China}}
\email{liuchao@jit.edu.cn}

\author{Tian Wang}
\affiliation{%
	\institution{Beihang University}
	\city{Beijing}
	\country{China}}
\email{wangtian@buaa.edu.cn}

\author{Yifeng Li}
\affiliation{%
	\institution{Nanjing Tech University}
	\city{Nanjing}
	\country{China}}
\email{lyffz4637@163.com}
\orcid{0000-0003-4798-3211}

\renewcommand{\shortauthors}{Zijie Wang et al.}

\begin{abstract}
  Given a natural language description, text-based person retrieval aims to identify images of a target person from a large-scale person image database. Existing methods generally face a \textbf{color over-reliance problem}, which means that the models rely heavily on color information when matching cross-modal data. Indeed, color information is an important decision-making accordance for retrieval, but the over-reliance on color would distract the model from other key clues (e.g. texture information, structural information, etc.), and thereby lead to a sub-optimal retrieval performance. To solve this problem, in this paper, we propose to \textbf{C}apture \textbf{A}ll-round \textbf{I}nformation \textbf{B}eyond \textbf{C}olor (\textbf{CAIBC}) via a jointly optimized multi-branch architecture for text-based person retrieval. CAIBC contains three branches including an RGB branch, a grayscale (GRS) branch and a color (CLR) branch. Besides, with the aim of making full use of all-round information in a balanced and effective way, a mutual learning mechanism is employed to enable the three branches which attend to varied aspects of information to communicate with and learn from each other. Extensive experimental analysis is carried out to evaluate our proposed CAIBC method on the CUHK-PEDES and RSTPReid datasets in both \textbf{supervised} and \textbf{weakly supervised} text-based person retrieval settings, which demonstrates that CAIBC significantly outperforms existing methods and achieves the state-of-the-art performance on all the three tasks.
\end{abstract}



\maketitle
\section{Introduction}

Given a natural language description, text-based person retrieval aims to identify images of a target person from a large-scale person image database. As in most real application scenarios, textual descriptions are much more accessible than other type of queries (e.g. images), text-based person retrieval is of significance in the field of video surveillance and has drawn more and more attention. Nevertheless, currently the majority of existing methods mainly focus on image-based person retrieval (aka. person re-identification), while the study of text-based person retrieval is still in its infancy.

\begin{figure}[!ht]
	\centering
	\includegraphics[width=\linewidth]{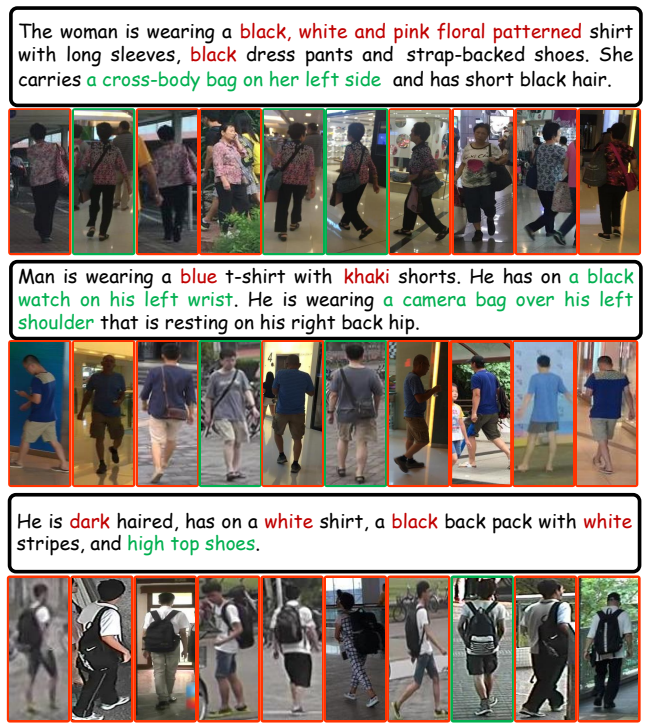}
	\caption{Text-based person retrieval examples given by a single-branch baseline model. The matched (mismatched) person images are marked with green (red) rectangles.}
	\label{fig:motivation-res}
\end{figure}

The core problem of text-based person retrieval is how to properly capture and match discriminative clues from multi-modal data, namely, raw person images and textual descriptions. Many approaches have been proposed to handle this problem. However, most of the existing methods are generally still faced with a common difficulty, which can be termed as the \textbf{color over-reliance problem}. Some typical incorrect text-based person retrieval examples given by a single-branch baseline model (detailed in Sec.~\ref{sec:baseline}) are illustrated in Fig.~\ref{fig:motivation-res}. The matched (mismatched) person images are marked with green (red) rectangles. It can be observed that many of the mismatched images in the retrieved top-10 list well meet the color information in the query sentence and have similar colors with the matched images. However, some tiny but discriminative clues like `cross-body bag', `camera bag', `high top shoes' in the examples are ignored by the model, which as a result leads to the failure in retrieval. And in some cases, even if the colors in an image are corresponding to wrong local parts according to the query sentence, the image still ranks high. All these observations suggest that existing methods rely heavily on color information when matching multi-modal data.

\begin{figure}[!ht]
	\centering
	\includegraphics[width=\linewidth]{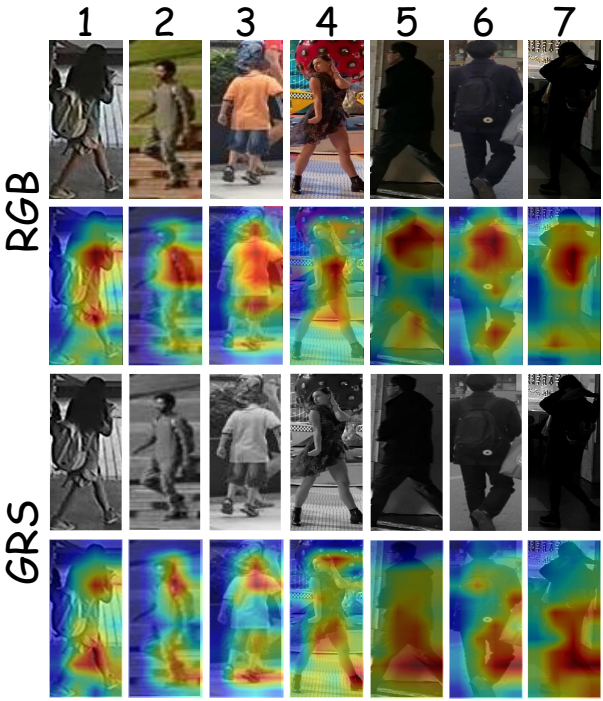}
	\caption{Feature response maps for RGB and grayscale person images, which are generated respectively by single-branch models trained on data with/without color information. The response map of each image is calculated by the mean of all feature maps. First and third rows denote the RGB and grayscale images, respectively, and second and fourth rows are their corresponding feature response maps.}
	\label{fig:motivation-atten}
\end{figure}

Indeed, color information is an important decision-making accordance for retrieval, but the over-reliance on color would distract the model from other key clues (e.g. texture information, structural information, etc.), and thereby leads to a sub-optimal retrieval performance. So it seems that managing to alleviate this \textbf{color over-reliance problem} can be a key to further promote the research on this task. In Fig.~\ref{fig:motivation-atten}, we display the feature response maps of several RGB/grayscale (GRS) image pairs, which are respectively generated by two single-branch baseline models trained on data with/without color information. It can be observed that for RGB and GRS data, the models focus on different local regions, and hence it is reasonable to deduce that the joint utilization of RGB and GRS data could benefit from the complementary effect between them and alleviate the color over-reliance problem.

To this end, in this paper, we propose to \textbf{C}apture \textbf{A}ll-round \textbf{I}nformation \textbf{B}eyond \textbf{C}olor (\textbf{CAIBC}) via a jointly optimized multi-branch architecture for text-based person retrieval. Specifically, besides a conventionally employed \textbf{RGB branch} which takes raw RGB images and description sentences as input, two more branches including a \textbf{grayscale (GRS) branch} and a \textbf{color (CLR) branch} are further proposed to handle the color over-reliance problem. For the GRS branch, a Color Deprivation Module (CDM) is utilized to deprive an input RGB image of color information and output a grayscale image, while a Color Masking Module (CMM) is adopted to mask all the color-related words in a textual description. After that, as the color information contained in multi-modal data is almost completely removed, there is no chance for the GRS branch to rely on color information when retrieving, and hence it is supposed to seek for other discriminative clues beyond color. In addition, as the goal of CAIBC is to make full use of all-round information in a balanced and effective way, rather than overly emphasize on some information and ignore the other, the situation that the model well captures non-color clues but lose attention on discriminative color information is also not desired. Therefore, a CLR branch is further employed to explicitly care for color information, in which a Color Prior Module (CPM) is employed to enhance the extracted color information. In addition, a mutual learning mechanism \cite{mutuallearning} is adopted to enable the three branches to communicate with and learn from each other. We evaluate our proposed method on the CUHK-PEDES \cite{Shuang2017Person} and RSTPReid \cite{dssl} datasets. Experimental results demonstrate that CAIBC outperforms previous methods and achieves the state-of-the-art performance in both \textbf{supervised} \cite{Shuang2017Person} and \textbf{weakly supervised} \cite{zhao2021CMMT} text-based person retrieval settings. As further discussed in Sec.~\ref{sec:compare-with-sota}, by capturing all-round information beyond color, CAIBC outperforms existing works on the weakly-supervised text-based person retrieval task without any clustering or pseudo label generating based methods.

The main contributions of this paper can be summarized as fourfold: 
\begin{itemize}
	\item A jointly optimized multi-branch architecture termed as \textbf{CAIBC} is proposed for text-based person retrieval to \textbf{C}apture \textbf{A}ll-round \textbf{I}nformation \textbf{B}eyond \textbf{C}olor and tackle the \textbf{color over-reliance problem}, which contains three branches including an RGB branch, a grayscale (GRS) branch and a color (CLR) branch.
	\item A \textbf{mutual learning} mechanism is employed to enable the three branches to communicate with and learn from each other, and hence to make full use of all-round information in a balanced and effective way.
	\item To our knowledge, we are the first to consider utilizing grayscale data along with RGB data to boost the performance on the task of Text-based Person Retrieval.
	\item Extensive experimental analysis demonstrates that CAIBC significantly outperforms existing methods and achieves the state-of-the-art performance in both \textbf{supervised} and \textbf{weakly supervised} text-based person retrieval settings.
\end{itemize}

\begin{figure*}[!ht]
	\centering
	\includegraphics[width=\linewidth]{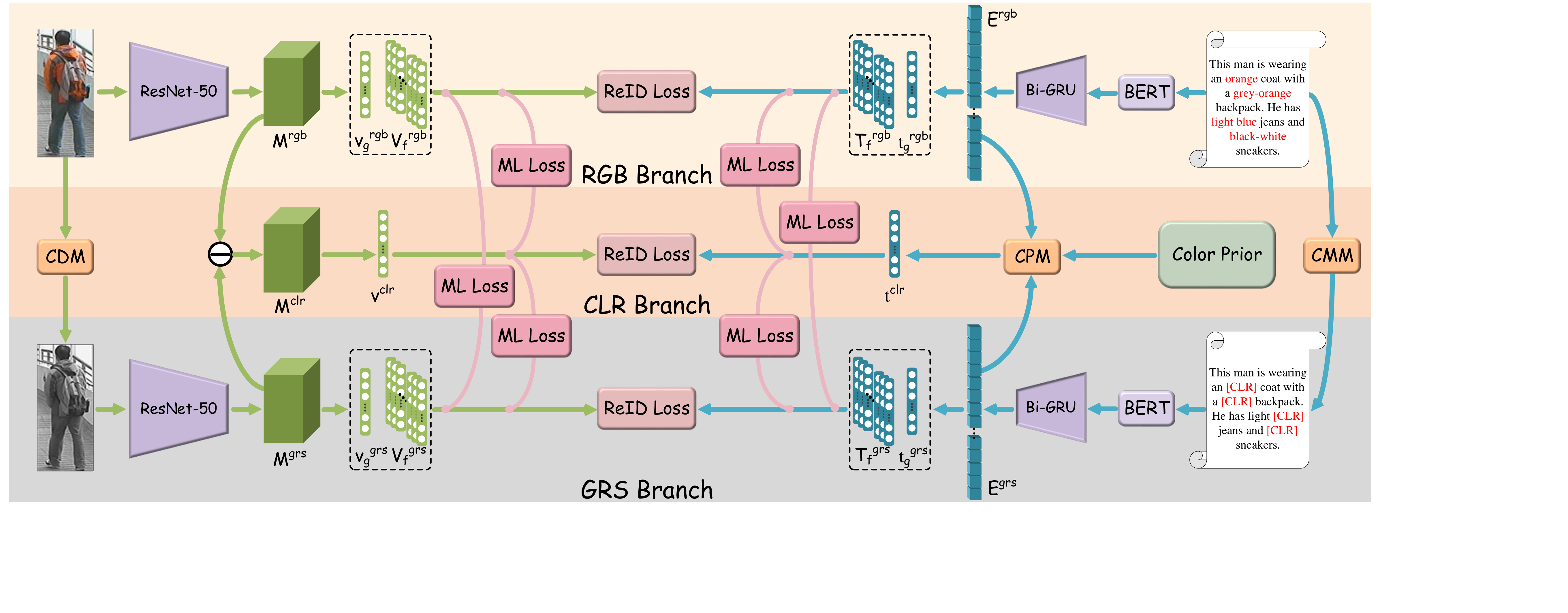}
	\caption{The overall framework of the proposed CAIBC model, which contains an RGB branch, a grayscale (GRS) branch and a color (CLR) branch to capture all-round information beyond color.}
	\label{fig:model}
\end{figure*}

\section{Related Works}
\subsection{Person Re-identification}
The goal of Person re-identification is to identify the targeted person images given a query image. The key challenges lie in the large intra-class and small inter-class variation caused by different views, poses, illuminations, and occlusions \cite{YSun,zhong2019invariance,SUM,sun2019perceive}. Yi et al. \cite{yi2014deepreid} firstly proposed deep learning methods to match people with the same identification. In order to strengthen the representation capability of the deep neural network, Hou et al. \cite{IAM2019CVPR} proposed the Interaction-and-Aggregation (IA) Block. Hao et al. \cite{hao2021crossYe} propose a Modality Confusion Learning Network (MCLNet), of which the basic idea is to confuse two modalities, ensuring that the optimization is explicitly concentrated on the modality-irrelevant perspective.

\subsection{Text-based Person Retrieval}
Text-based person retrieval aims to search for the corresponding pedestrian image according to a given text query. This task is first introduced by Li et al. \cite{Shuang2017Person} and a GNA-RNN model is employed to handle the cross-modal data. Jing et al. \cite{jing2018pose} utilize pose information as soft attention to localize the discriminative regions. Niu et al. \cite{niu2020improving} adopt a Multi-granularity Image-text Alignments (MIA) model exploit the combination of multiple granularities. Liu et al. \cite{mm2019graphreid} generate fine-grained structured representations from images and texts of pedestrians with an A-GANet model to exploit semantic scene graphs. CMAAM \cite{aggarwal2020text} is proposed to learn an attribute-driven space along with a class-information driven space by introducing extra attribute annotation and prediction. Zheng et al. \cite{zheng2020hierarchical} propose a Gumbel attention module to alleviate the matching redundancy problem and a hierarchical adaptive matching model is employed to learn subtle feature representations from three different granularities. Zhu er al. \cite{dssl} proposed a Deep Surroundings-person Separation Learning (DSSL) model to effectively extract and match person information. Besides, they construct a Real Scenarios Text-based Person Re-identification (RSTPReid) dataset to benefit future research. Wu et al. \cite{wu2021lapscore} introduce two color reasoning sub-tasks including image colorization and text completion to enable the model to learn fine-grained cross-modal association. Zhao et al. \cite{zhao2021CMMT} introduce the weakly supervised person retrieval task and proposed a Cross-Modal Mutual Training (CMMT) framework to handle this problem. Jing et al. \cite{jing2020MAN} combines the challenges from both cross-modal (text-based) person search and cross-domain person search, and a moment alignment network (MAN) to solve this task.

\section{Methodology}
\label{sec:method}

\subsection{Problem Formulation}
The goal of the proposed framework (shown in Fig.~\ref{fig:model}) is to measure the similarity between cross-modal data, namely, a given textual description query and a gallery person image. Formally, let $D = \{p_{i}, q_{i}\}_{i=1}^{N}$ denotes a dataset consists of $N$ image-text pairs. Each pair contains a pedestrian image $p_{i}$ captured by one certain surveillance camera along with its corresponding textual description query $q_{i}$. The IDs of the $Q$ varied pedestrians in the dataset are denoted as $Y = \{y_{i}\}_{i=1}^{N}$ with $y_{i} \in \{1, \cdots, Q\}$. Given a textual description, the aim is to identify images of the most relevant pedestrian from a large scale person image gallery.

\subsection{Branch Prototype}
\label{sec:baseline}
First, we introduce the branch prototype which extracts and matches multi-granular representations from multi-modal data. It can be utilized to implement both the RGB and GRS branches in CAIBC.
\subsubsection{\textbf{Visual Representation Extraction}}
Given either an RGB or grayscale image, a pretrained ResNet-50 \cite{ResNet} backbone is utilized to extract multi-granular visual representations. To obtain the global representation $v_{g} \in \mathbb{R}^{P}$, the feature map before the last pooling layer of ResNet-50 is down-scaled with a global max pooling (GMP) operation and converted to a $2048$-dim vector, which is then passed through a fully-connected (FC) layer and transformed to $P$-dim. In the local branch, the same feature map is first horizontally $k$-partitioned with GMP, and then the local strips are separately passed through an FC layer to form $K$ $P$-dim fine-grained local visual representations $V_{f} = \{v_{fk}\}_{k=1}^{K}$.
\subsubsection{\textbf{Textual Representation Extraction}}
For textual representation extraction, an input sentence is processed by a bi-directional Gated Recurrent Unit (bi-GRU) after the words are embedded via a pretrained BERT language model \cite{vaswani2017BERT}. The $i$-th last hidden states of the forward and backward GRUs are averaged to represent the $i$-th word in the query sentence as $e_{i} \in \mathbb{R}^C$. The overall $n$ word representations are concatenated as $E \in \mathbb{R}^{n \times C}$ to represent the whole sentence, on which a row-wise max pooling (RMP) operation followed by an FC layer is performed to get the global textual representation $t_{g} \in \mathbb{R}^{P}$. After that, we propose a Word Attention Module following SSAN \cite{ding2021ssan} to obtain $K$ local textual representations according to word-part correspondences:
\begin{equation}
	s_{i}^{k} = \sigma(W_{p}^{k}e_{i}), \ E_{k} = \{s_{i}^k e_{i}\}_{i=1}^{n}, \ k = \{1, 2, \dots, K\},
\end{equation}
where $\sigma$ denotes the Sigmoid function and $W_{p}^{k} \in \mathbb{R}^{1 \times C}$ stands for a linear transformation operation. Then each modified sentence representation $E_{k}$ is processed separately by RMP + FC and stacked to obtain the fine-grained local textual representations $T_{f} = \{t_{fk}\}_{k=1}^{K}$.

After obtaining both visual and textual multi-granular representations, the fine-grained feature representations in $V_{f}$ and $T_{f}$ are first concatenated with each other to form the unified visual and textual local feature representations $v_{l}$ and $t_{l}$ $\in \mathbb{R}^{KP}$. And then the cross-modal global and local similarities can be calculated as
\begin{equation}
	S_{g} = \frac{v_{g}^{T} t_{g}}{||v_{g}|| ||t_{g}||}, \ S_{l} = \frac{v_{l}^{T} t_{l}}{||v_{l}|| ||t_{l}||}.
\end{equation}

\subsection{Color Deprivation and Masking}
Given a certain RGB image $\in \mathbb{R}^{3 \times H \times W}$, its corresponding grayscale image can be conveniently obtained via a \textbf{Color Deprivation Module (CDM)}, which can be formulated as:
\begin{equation}
	GRS(i, j) = \begin{bmatrix}
		0.299 & 0.587 & 0.114\\
	\end{bmatrix}
	\begin{bmatrix}
		R(i, j) \\
		G(i, j) \\
		B(i, j) \\
	\end{bmatrix},
\end{equation}
where $i \in \{1, 2, \dots, H\}, \ j \in \{1, 2, \dots, W\}$. $GRS(i, j)$, $R(i, j)$, $G(i, j)$ and $B(i, j)$ denote the pixel value in the $i$-th row and the $j$-th column of the grayscale image and the three RGB channels, respectively. 0.299, 0.587 and 0.114 are empirical coefficients for RGB-to-GRS conversion which are widely used in famous tools like OpenCV and Photoshop. To process the obtained grayscale images with ResNet-50, we expend them to three channels by duplicating the grayscale channel three times.

\begin{figure}[h]
	\centering
	\includegraphics[width=\linewidth]{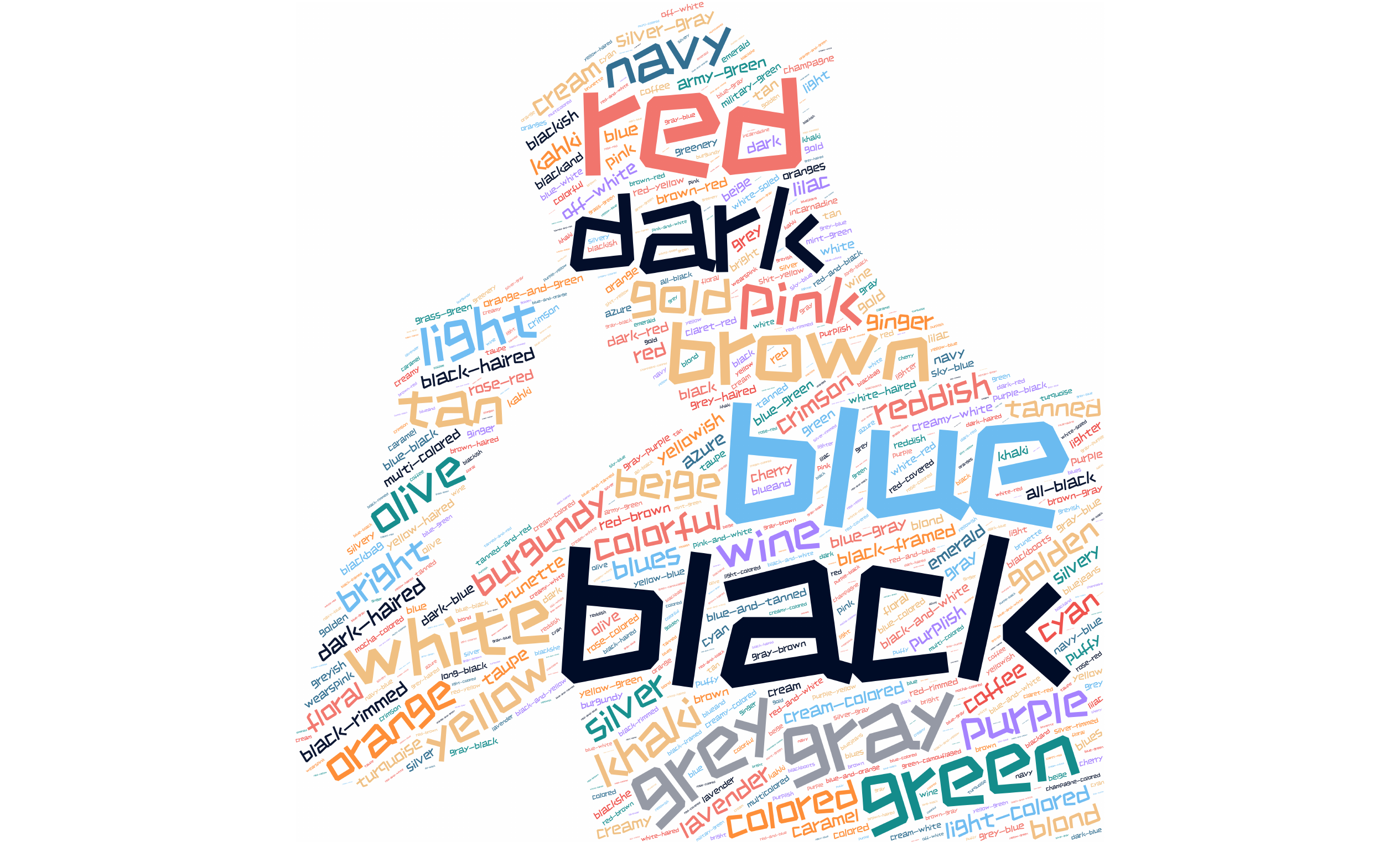}
	\caption{Word cloud constructed based on the frequency of color-related words in the color bank.}
	\label{fig:cloud}
	\vspace{-0.6cm}
\end{figure}

When it comes to color removing for the textual modality, a \textbf{Color Masking Module (CMM)} is proposed. First, with the person description corpus from text-based person retrieval datasets, a \textbf{color bank} is constructed by collecting color-related words that appear with high frequency. The word cloud constructed based on the frequency of color-related words is illustrated in Fig.~\ref{fig:cloud}. And then given an input textual description, all the color-related words are masked as a unified token $[CLR]$ to remove the color information.

Note that both CDM and CMM \textit{have no parameters to learn} and can be simply applied to the input raw multi-modal data.

\subsection{Capturing All-round Information via Multi-branch Architecture}

\subsubsection{\textbf{RGB Branch}}
The RGB branch takes an RGB image along with a textual description as input, which outputs visual/textual multi-modal RGB representations as $v_{g}^{rgb}$ / $t_{g}^{rgb}$ $\in \mathbb{R}^{P}$ and $v_{l}^{rgb}$ / $t_{l}^{rgb}$ $\in \mathbb{R}^{KP}$. The similarities for this branch are computed as
\begin{equation}
	S_{g}^{rgb} = \frac{(v_{g}^{rgb})^{T} t_{g}^{rgb}}{||v_{g}^{rgb}|| ||t_{g}^{rgb}||}, \ S_{l}^{rgb} = \frac{(v_{l}^{rgb})^{T} t_{l}^{rgb}}{||v_{l}^{rgb}|| ||t_{l}^{rgb}||}.
\end{equation}
\subsubsection{\textbf{GRS Branch}}
The GRS branch takes in a converted grayscale image and a textual description with color-related words masked. Similar to the RGB branch, it gives out $v_{g}^{grs}$ / $t_{g}^{grs}$ $\in \mathbb{R}^{P}$ along with $v_{l}^{grs}$ / $t_{l}^{grs}$ $\in \mathbb{R}^{KP}$, and the similarities for the GRS branch are
\begin{equation}
	S_{g}^{grs} = \frac{(v_{g}^{grs})^{T} t_{g}^{grs}}{||v_{g}^{grs}|| ||t_{g}^{grs}||}, \ S_{l}^{grs} = \frac{(v_{l}^{grs})^{T} t_{l}^{grs}}{||v_{l}^{grs}|| ||t_{l}^{grs}||}.
\end{equation}
\subsubsection{\textbf{CLR Branch}}
As discussed in the Introduction, the goal of CAIBC is to make full use of all-round information in a balanced and effective way, rather than overly emphasize on some information and ignore the other, so the situation that the model well captures non-color clues but lose attention on discriminative color information is also not desired. To this end, we further propose the color (CLR) branch to explicitly care for color information and hence give a more stable and robust retrieval performance. Intuitively, as the RGB branch cares for general information while the GRS branch cares for discriminative clues beyond color, the non-shared information between the two branches can be pure color information. For the visual modality, the output feature map $M^{rgb}$ of ResNet-50 in the RGB branch is subtracted by $M^{grs}$ from the GRS branch to obtain a pure color feature map $M^{clr}$, which is then processed by GMP along with an FC layer to give the visual color representation $v^{clr} \in \mathbb{R}^{P}$. For the textual modality, a \textbf{Color Prior Module (CPM)} is proposed. Specifically, the representation $E^{rgb}$ of a whole sentence is first subtracted by the masked sentence representation $E^{grs}$ to get $E^{clr}$. And then the color-related words in the sentence are utilized as \textbf{color prior} to further enhance the color information contained in $E^{clr}$. To be specific, after word embedding, the color prior are summed up and converted as the same dimension with the word representation $e^{clr}_{i}$ in $E^{clr}$, which is then added with each $e^{clr}_{i}$ to give a color prior enhanced representation $E^{cp}$. With an RMP operation and an FC layer, the textual color representation $t^{clr} \in \mathbb{R}^{P}$ can be obtained. Eventually, the similarity for the CLR branch is calculated as
\begin{equation}
	S^{clr} = \frac{(v^{clr})^{T} t^{clr}}{||v^{clr}|| ||t^{clr}||}.
\end{equation}

\subsection{Optimization}

As discussed above, the core target of CAIBC is to synchronously learn the three branches and make full use of all-round information in a balanced and effective way. Thus a \textbf{mutual learning} mechanism is employed to enable knowledge communication across branches. First, the Kullback Leibler (KL) Divergence is utilized to quantify the similarity of logits from different branches. Let $v^{br}_{i} \in \{v^{rgb}_{gi}, v^{grs}_{gi}, v^{clr}_{i}\}$ and $t^{br}_{i} \in \{t^{rgb}_{gi}, t^{grs}_{gi}, t^{clr}_{i}\}$ denote the visual and textual features extracted from the $i$-th sample pair, which are used to calculated the probabilities $p^{br}_{vm}$ and $p^{br}_{tm}$ of class (person ID) $m$ as:
\begin{equation}
	p^{br}_{vm}(v^{br}_{i}) = \frac{exp(\gamma W_{m} v^{br}_{i})}{\sum_{k=1}^{M} exp(\gamma W_{k} v^{br}_{i})}, 
	p^{br}_{tm}(t^{br}_{i}) = \frac{exp(\gamma W_{m} t^{br}_{i})}{\sum_{k=1}^{M} exp(\gamma W_{k} t^{br}_{i})}, 
\end{equation}
where $\gamma W_{m} v^{br}_{i}$ and $\gamma W_{m} t^{br}_{i}$ are logits fed into the softmax layer in the $br$ branch ($br$ can be $rgb$, $grs$ or $clr$). And then each branch is optimized under the constraint of a \textbf{mutual learning (ML) loss}, which can be formulated as:
\begin{multline}
	\mathcal{L}_{ML}^{br} = \frac{1}{2} (\sum_{s \in B / \{br\}} \sum_{i=1}^{N} \sum_{m=1}^{M} p_{vm}^{br}(v^{br}_{i}) log \frac{p_{vm}^{br}(v^{br}_{i})}{p_{vm}^{s}(v^{br}_{i})} \\
	+ \sum_{s \in B / \{br\}} \sum_{i=1}^{N} \sum_{m=1}^{M} p_{tm}^{br}(t^{br}_{i}) log \frac{p_{tm}^{br}(t^{br}_{i})}{p_{tm}^{s}(t^{br}_{i})}),
\end{multline}
where $B = \{rgb, grs, clr\}$.

Besides, the commonly utilized triplet ranking loss along with identification (ID) loss are also adopted to train CAIBC, which are together denoted as \textbf{ReID loss}. Note that for local representations, the ID loss is imposed on each $v_{fk}^{rgb}$ / $t_{fk}^{rgb}$ and $v_{fk}^{grs}$ / $t_{fk}^{grs}$, while the triplet ranking loss is imposed on $v_{l}^{rgb}$ / $t_{l}^{rgb}$ and $v_{l}^{grs}$ / $t_{l}^{grs}$.

\section{Experiments}

\subsection{Experimental Setup}

\subsubsection{\textbf{Dataset}}
Our approach is evaluated on two challenging Text-based Person Retrieval datasets : CUHK-PEDES \cite{Shuang2017Person} and RSTPReid \cite{dssl}. (1) \textbf{CUHK-PEDES}: Following the official data split approach \cite{Shuang2017Person}, the training set of CUHK-PEDES contains 34054 images, 11003 persons and 68126 textual descriptions. The validation set contains 3078 images, 1000 persons and 6158 textual descriptions while the test set has 3074 images, 1000 persons and 6156 descriptions. (2) \textbf{RSTPReid}: The RSTPReid dataset contains 20505 images of 4,101 persons. Each person has 5 corresponding images taken by different cameras and each image is annotated with 2 textual descriptions. For data division, 3701, 200 and 200 identities are utilized for training, validation and test, respectively.
\subsubsection{\textbf{Evaluation Metrics}}
The performance is evaluated by the Rank-k accuracy (R@k). All images in the test set are ranked by their similarities with a given query natural language sentence. If any image of the corresponding person is contained in the top-k images, we call this a successful search. We report the Rank-1/5/10 accuracies for all experiments.

\begin{table}[!ht]
	\caption{Comparison with SOTA (supervised) on CUHK-PEDES.}
	\label{tab:Sota-CUHK}
	\begin{tabular}{l|ccc}
		\toprule
		Method & R@1 & R@5 & R@10 \\
		\midrule
		CNN-RNN \cite{reed2016learning}& 8.07  & -  & 32.47 \\
		Neural Talk \cite{vinyals2015show} & 13.66  & -  & 41.72 \\
		GNA-RNN \cite{Shuang2017Person} & 19.05  & -  & 53.64 \\
		IATV \cite{li2017identity} & 25.94  & -  & 60.48 \\
		PWM-ATH \cite{Chen2018} & 27.14  & 49.45  & 61.02 \\
		Dual Path \cite{zheng2020dual} & 44.40  & 66.26  & 75.07 \\
		GLA \cite{chen2018improving} & 43.58  & 66.93  & 76.26 \\
		CMPM-CMPC \cite{zhang2018CMPM-CMPC} & 49.37 & 71.69 & 79.27 \\
		MIA \cite{niu2020improving} & 53.10 & 75.00 & 82.90 \\
		A-GANet \cite{mm2019graphreid} & 53.14 & 74.03 & 81.95 \\
		PMA \cite{jing2018pose} & 54.12 & 75.45 & 82.97 \\
		TIMAM \cite{ARL} & 54.51 & 77.56 & 84.78 \\
		ViTAA \cite{wang2020vitaa} & 55.97 & 75.84 & 83.52 \\
		CMAAM \cite{aggarwal2020text} & 56.68 &	77.18 &	84.86 \\
		HGAN \cite{zheng2020hierarchical} & 59.00 & 79.49 & 86.62 \\
		NAFS \cite{gao2021contextual} & 59.94 & 79.86 & 86.70 \\
		DSSL \cite{dssl} & 59.98 & 80.41 & 87.56 \\
		MGEL \cite{wang2021ijcai} & 60.27 & 80.01 & 86.74 \\
		SSAN \cite{ding2021ssan} & 61.37 & 80.15 & 86.73 \\
		NAFS \cite{gao2021contextual} + TC\&IC \cite{wu2021lapscore} & \underline{63.40} & - & \underline{87.80} \\
		\midrule
		\textbf{CAIBC w/o BERT (Ours)} & 62.33 & \underline{81.32} & 87.35  \\
		\textbf{CAIBC (Ours)} & \textbf{64.43} & \textbf{82.87} & \textbf{88.37} \\
		\bottomrule
	\end{tabular}
\end{table}

\begin{table}[!ht]
	\caption{Comparison with SOTA on RSTPReid.}
	\label{tab:Sota-RSTPReid}
	\begin{tabular}{l|ccc}
		\toprule
		Method & R@1 & R@5 & R@10 \\
		\midrule
		IMG-Net \cite{wang2020img} & 37.60 & 61.15 & 73.55 \\
		AMEN \cite{wang2021amen} & 38.45 & 62.40 & 73.80 \\
		DSSL \cite{dssl} & 39.05 & 62.60 & 73.95 \\
		SSAN \cite{ding2021ssan} & 43.50 & 67.80 & 77.15 \\
		\midrule
		\textbf{CAIBC w/o BERT (Ours)} & \underline{45.05} & \underline{68.15} & \underline{77.65} \\
		\textbf{CAIBC (Ours)} & \textbf{47.35} & \textbf{69.55} & \textbf{79.00} \\
		\bottomrule
	\end{tabular}
\end{table}

\begin{table}[!ht]
	\caption{Comparison with SOTA (weakly supervised) on CUHK-PEDES.}
	\label{tab:Sota-CUHK-ws}
	\begin{tabular}{l|ccc}
		\toprule
		Method & R@1 & R@5 & R@10 \\
		\midrule
		MM-TIM \cite{gomez2019MM-TIM} & 45.35 & 63.78 & 70.63 \\
		CMPM + MMT \cite{ge2020MMT} & 50.51 & 70.23 & 78.98 \\
		CMPM + SpCL \cite{SpCL} & 51.13 & 71.54 & 80.03 \\
		CMMT \cite{zhao2021CMMT} & 57.10 & 78.14 & \underline{85.23} \\
		\midrule
		\textbf{CAIBC w/o BERT (Ours)} & \underline{57.18} & \underline{78.17} & 85.15  \\
		\textbf{CAIBC (Ours)} & \textbf{58.64} & \textbf{79.02} & \textbf{85.93}\\
		\bottomrule
	\end{tabular}
\end{table}

\subsection{Implementation Details}
In our experiments, we set the representation dimensionality $P = 2048$. The dimensionality of embedded word vectors is set to 500. Two separate ResNet-50 \cite{ResNet} models pretrained on the ImageNet database \cite{russakovsky2015imagenet} are utilized as the visual CNN backbone for the RGB and GRS branch and a pretrained BERT language model \cite{vaswani2017BERT} is used to handle the textual input. For the CUHK-PEDES and RSTPReid dataset, the input images are resized to $384 \times 128 \times 3$. $K$ is set to 6. The random horizontal flipping strategy is employed for data augmentation. An Adam optimizer \cite{AdamOptimizer} is adopted to train the model with a batch size of 64. The parameters of BERT is fixed during training. The learning rate is initially set as 0.0001 for fine-tuning the two pretrained ResNet-50 backbones and 0.001 for the rest parts of our proposed model. The number of training epochs is set to 100.

\subsection{Results}

\subsubsection{\textbf{Comparison with SOTA in Supervised Settings}}
We compare our proposed CAIBC method with previous approaches in the supervised setting \cite{Shuang2017Person} on CUHK-PEDES and RSTPReid, as shown in Tab.~\ref{tab:Sota-CUHK} and Tab.~\ref{tab:Sota-RSTPReid}, respectively. Note that in all the tables of this paper, the highest values are marked as bold type while the second high values are underlined. It can be observed that by capturing all-round information beyond color, CAIBC achieves the state-of-the-art performance on both datasets. Specifically, Lapscore \cite{wu2021lapscore} introduces two color reasoning sub-tasks including text completion and image colorization (TC$\&$IC) during training to enable the model to better excavate color information and learn fine-grained cross-modal association. On the contrary, the goal of our proposed CAIBC method is to enable the model to take full advantage of other key clues along with color, so as to alleviate the color over-reliance problem. Besides, the two sub-tasks proposed in Lapscore can also be employed as add-on modules in CAIBC to further improve the performance, which remains our future work in the extension version.

\subsubsection{\textbf{Comparison with SOTA in Weakly Supervised Settings}}
In addition, we also evaluate CAIBC on the weakly supervised person retrieval task \cite{zhao2021CMMT} and display the comparison with previous methods in Tab.~\ref{tab:Sota-CUHK-ws}. It can be noticed that CAIBC achieves competitive and even slightly better performance compared with existing SOTA methods on the weakly-supervised text-based person retrieval task without any clustering or pseudo label generating based methods (further discussed in Sec.~\ref{sec:compare-with-sota}), which indicates that our proposed method indeed alleviate the color over-reliance problem to some extent. 

\begin{table*}[!ht]
	\caption{Ablation analysis of key components on CUHK-PEDES and RSTPReid. $\mathcal{CP}$ stands for the utilization of color prior.}
	\label{tab:abla}
	\begin{tabular}{c|ccc|ccccc|ccc|ccc}
		\toprule
		No. & \multicolumn{3}{c|}{Branches} & \multicolumn{5}{c|}{Components} & \multicolumn{3}{c|}{CUHK-PEDES} & \multicolumn{3}{c}{RSTPReid}\\
		\midrule
		- & RGB & GRS & CLR & $\mathcal{CP}$ & $\mathcal{L}_{ML}$ & $\mathcal{L}_{id}$ & $\mathcal{L}_{tri}$ & BERT & R@1 & R@5 & R@10 & R@1 & R@5 & R@10 \\
		\midrule
		Exp.1 & $\checkmark$ & $\times$ & $\times$ & $\times$ & $\times$ & $\checkmark$ & $\checkmark$ & $\times$ & 58.97 & 79.03 & 85.38 & 40.05 & 64.85 & 75.20 \\
		Exp.2 & $\times$ & $\checkmark$ & $\times$ & $\times$ & $\times$ & $\checkmark$ & $\checkmark$ & $\times$ & 24.89 & 46.43 & 57.02 & 13.45 & 32.00 & 44.25 \\
		\midrule
		Exp.3 & $\checkmark$ & $\mathcal{T}$ & $\times$ & $\times$ & $\times$ & $\checkmark$ & $\checkmark$ & $\times$ & 59.68 & 79.22 & 86.21 & 40.15 & 65.30 & 75.60 \\
		Exp.4 & $\checkmark$ & $\mathcal{V}$ & $\times$ & $\times$ & $\times$ & $\checkmark$ & $\checkmark$ & $\times$ & 59.97 & 79.40 & 86.34 & 41.85 & 66.85 & 77.00 \\
		Exp.5 & $\checkmark$ & $\checkmark$ & $\times$ & $\times$ & $\times$ & $\checkmark$ & $\checkmark$ & $\times$ & 60.10 & 79.91 & 86.34 & 42.15 & 67.10 & 76.75 \\
		Exp.6 & $\checkmark$ & $\checkmark$ & $\times$ & $\times$ & $\checkmark$ & $\checkmark$ & $\checkmark$ & $\times$ & 61.01 & 80.33 & 86.71 & 43.10 & 67.45 & 77.15 \\
		\midrule
		Exp.7 & $\checkmark$ & $\checkmark$ & $\checkmark$ & $\times$ & $\times$ & $\checkmark$ & $\checkmark$ & $\times$ & 61.03 & 80.46 & 87.20 & 42.65 & 67.05 & 76.95 \\
		Exp.8 & $\checkmark$ & $\checkmark$ & $\checkmark$ & $\checkmark$ & $\times$ & $\checkmark$ & $\checkmark$ & $\times$ & 61.68 & 80.85 & 87.18 & 43.70 & 67.35 & 77.05 \\
		Exp.9 & $\checkmark$ & $\checkmark$ & $\checkmark$ & $\times$ & $\checkmark$ & $\checkmark$ & $\checkmark$ & $\times$ & 62.02 & 80.73 & 87.26 & 44.25 & 67.70 & 77.35 \\
		\midrule
		Exp.10 & $\checkmark$ & $\times$ & $\times$ & $\times$ & $\times$ & $\times$ & $\checkmark$ & $\times$ &  53.64 & 74.66 & 82.44 & 36.30 & 61.10 & 71.75 \\
		Exp.11 & $\times$ & $\checkmark$ & $\times$ & $\times$ & $\times$ & $\times$ & $\checkmark$ & $\times$ & 21.30 & 42.48 & 53.72 & 10.10 & 25.05 & 37.10 \\
		Exp.12 & $\checkmark$ & $\checkmark$ & $\times$ & $\times$ & $\times$ & $\times$ & $\checkmark$ & $\times$ & 55.46 & 76.41 & 83.80 & 39.05 & 62.95 & 74.40 \\
		Exp.13 & $\checkmark$ & $\checkmark$ & $\times$ & $\times$ & $\checkmark$ & $\times$ & $\checkmark$ & $\times$ & 56.48 & 77.10 & 84.19 & 39.40 & 63.80 & 74.95 \\
		Exp.14 & $\checkmark$ & $\checkmark$ & $\checkmark$ & $\checkmark$ & $\times$ & $\times$ & $\checkmark$ & $\times$ & 56.60 & 77.26 & 84.52 & 40.20 & 64.85 & 75.40 \\
		Exp.15 & $\checkmark$ & $\checkmark$ & $\checkmark$ & $\checkmark$ & $\checkmark$ & $\times$ & $\checkmark$ & $\times$ & 57.18 & 78.17 & 85.15 & 40.45 & 65.20 & 75.75 \\
		Exp.16 & $\checkmark$ & $\checkmark$ & $\checkmark$ & $\checkmark$ & $\checkmark$ & $\times$ & $\checkmark$ & $\checkmark$ & 58.64 & 79.02 & 85.93 & 42.10 & 66.25 & 76.65 \\
		\midrule
		Exp.17 & $\checkmark$ & $\checkmark$ & $\checkmark$ & $\checkmark$ & $\checkmark$ & $\checkmark$ & $\checkmark$ & $\times$ & \underline{62.33} & \underline{81.32} & \underline{87.35} & \underline{45.05} & \underline{68.15} & \underline{77.65} \\
		Exp.18 & $\checkmark$ & $\checkmark$ & $\checkmark$ & $\checkmark$ & $\checkmark$ & $\checkmark$ & $\checkmark$ & $\checkmark$ & \textbf{64.43} & \textbf{82.87} & \textbf{88.37} & \textbf{47.35} & \textbf{69.55} & \textbf{79.00} \\
		\bottomrule
	\end{tabular}
\end{table*}

 \subsubsection{\textbf{Discussion on Superiority of CAIBC for Alleviating Color Over-reliance Problem}}
\label{sec:compare-with-sota}
In the context of the traditional supervised text-based person retrieval task, it seems that the negative effects caused by the color over-reliance problem may be slightly alleviated with the identity annotation for each pair as prior knowledge. To be specific, when the model is suffering from the color over-reliance problem and fails to properly distinguish person images with similar color information, the identity prior information can tell the model whether the two samples similar in color are belong to a same person or not. Nevertheless, depending all on the identity prior to alleviate the color over-reliance problem is just not enough. During the training process, all the identity prior can do is to tell the model that it has made a mistake in distinguishing samples with similar color information, but it is not able to tell the model exactly what details it has ignored, and hence the positive impact it can bring is quite limited.

Besides, when there is no identity information given and only the pairwise relationship is available (termed as Weakly Supervised Text-based Person ReID in \cite{zhao2021CMMT}), the impact of color over-reliance problem can become more significant. Many of the previous works propose clustering or pseudo label generating based methods for the weakly supervised text-based person retrieval task, with the aim to excavate some other prior knowledge to compensate for the lack of identity information. However, as these methods still depend on some kind of prior knowledge and most of the adopted priors are less precise than the directly annotated identity labels, there is no chance for them to better alleviate the color over-reliance problem.

In conclusion, as discussed above, most of the existing methods for either supervised and weakly supervised text-based person retrieval rely on some certain kind of prior knowledge to tell the model whether it is hindered by the color over-reliance problem without further detailed guidance. This kind of paradigms is indirect and thereby can only lead to a sub-optimal retrieval performance. Instead, within our proposed CAIBC method, the multi-branch architecture is able to explicitly care for different kinds of information. To be specific, the grayscale (GRS) branch handles multi-modal data with color information removed and is thereby forced to focus on other discriminative clues beyond color. On the contrary, the color (CLR) branch is proposed to explicitly attend to color information. In addition, the commonly utilized RGB branch is adopted to generally handle the raw input data. The three branches in CAIBC play varied roles, which complement and constrain each other when extracting and matching information from multi-modal data. Considering that each of the three proposed branches can be guided and corrected by the other two branches by means of the complementary effect among them, the reliance on prior knowledge to tackling the color over-reliance problem can be significantly released. In other words, when one certain branch neglect some key clues, the other branches are able to catch the missed information and still make the complete CAIBC model itself focus on sufficient discriminative clues, instead of waiting for the identity prior to tell it a vague true or false without explicit guidance.

\begin{figure*}[!ht]
	\centering
	\includegraphics[width=0.85\linewidth]{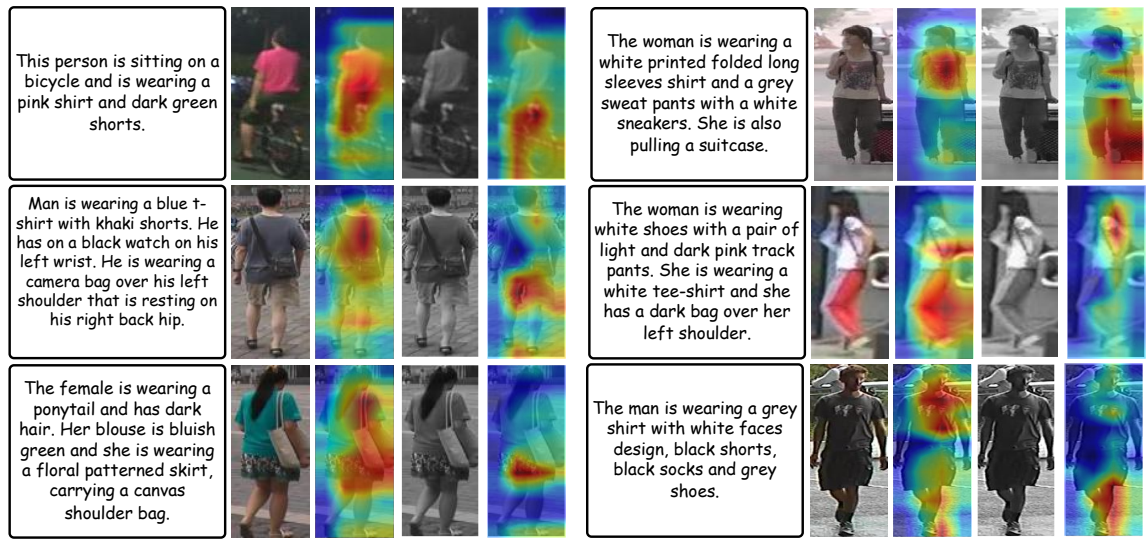}
	\caption{Visualization of Feature Response Maps within RGB and GRS Branches of CAIBC. For each example, the 1st and the 3rd images are respectively the RGB and GRS images fed into the RGB and GRS branches, while the 2nd and the 4th images are their corresponding feature response maps. Each feature response map is calculated by the mean of all feature maps}
	\label{fig:atten-main}
	\vspace{-0.1cm}
\end{figure*}

\subsection{Ablation Analysis}
To demonstrate the effectiveness and contribution of each component in CAIBC, a series of ablation experiments are carried out on CUHK-PEDES and RSTPReid. The experimental results are reported in Tab.~\ref{tab:abla}, which are numbered from 1 to 18. Exp.1 is conducted with a single RGB branch baseline model (detailed in Sec.~\ref{sec:baseline}) while Exp.2 is carried out with a single GRS branch model. Exp.3$\sim$6 are conducted with double branch structures and $\mathcal{V}$ or $\mathcal{T}$ in the table means that the GRS branch is employed only for the visual or textual modality. Exp.7$\sim$9 are performed with all the 3 branches while Exp.17 and Exp.18 is carried out with the complete CAIBC. From Exp.10 to Exp.16, we further carried out experimental analysis to see the effectiveness of our proposed components on weakly supervised person retrieval.
\subsubsection{\textbf{Impact of Multi-branch Learning}}
It can be observed from Tab.~\ref{tab:abla} that although there exists a performance gap between the model with a single GRS branch and the one with a single RGB branch, by combining the RGB branch with GRS branch, an obvious performance gain is obtained. And when the CLR branch which explicitly cares for color information is added, the retrieval accuracy is further improved. All these observations indicate that by means of the jointly optimized multi-branch architecture, CAIBC is enabled to separately care for different types of information from varied aspects and the 3 branches can complement each other to achieve a superior retrieval performance.
\subsubsection{\textbf{Impact of Mutual Learning}}
By comparing Exp.5 with Exp.6 in Tab.~\ref{tab:abla}, it can be observed that after the mutual learning mechanism is utilized, the performance increases by $0.91\%$, $0.42\%$, $0.37\%$ and $0.95\%$, $0.35\%$ $0.40\%$ on CUHK-PEDES and RSTPReid under the Rank-1/5/10 accuracies, respectively. Besides, comparing Exp.17 with Exp.8, the performance increases by $0.65\%$, $0.47\%$, $0.17\%$ and $1.30\%$, $0.80\%$ $0.60\%$ on CUHK-PEDES and RSTPReid, which proves the benefit of adopting the mutual learning mechanism to enable knowledge communication among branches.
\subsubsection{\textbf{Impact of Color Prior ($\mathcal{CP}$)}}
In addition, as can be seen from the two pairs of comparison (Exp.7, Exp.8) and (Exp.9, Exp.17), introducing color prior information into the model is able to further bring a performance gain, which proves its effectiveness to enhance the CLR branch.

\begin{figure}[!ht]
	\centering
	\includegraphics[width=\linewidth]{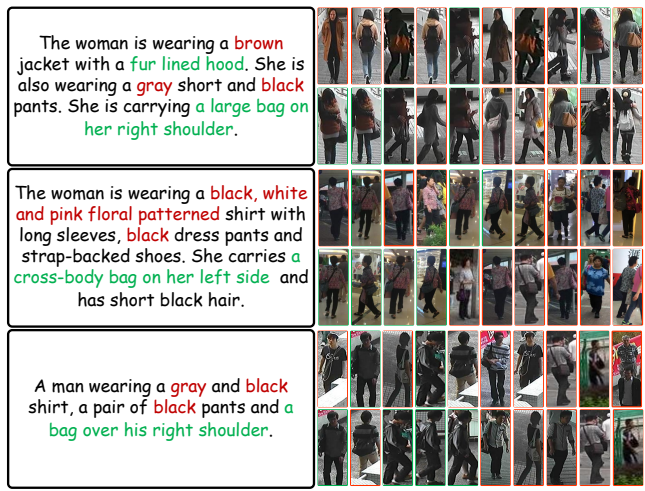}
	\caption{Text-based person retrieval examples given by CAIBC. Note that for each example, the first and the second row are given by a single-branch model trained on data with and without color information. The matched and mismatched person images are marked with green and red rectangles, respectively.}
	\label{fig:result}
 	\vspace{-0.4cm}
\end{figure}

\subsection{Visualization}
Some typical feature response maps within RGB and GRS Branches of CAIBC are visualized in Fig.~\ref{fig:atten-main}, while some of the text-based person retrieval examples are displayed in Fig.~\ref{fig:result}.

\section{Conclusion}
Text-based person retrieval aims to identify images of a target person from a large-scale person image database according a given natural language description. Existing methods still generally face a \textbf{color over-reliance problem}, which means the models rely heavily on color information when matching cross-modal data. Indeed, color information is an important decision-making accordance for retrieval, but the over-reliance on color would distract the model from other key clues (e.g. texture information, structure information, etc.), and thereby lead to a sub-optimal retrieval performance. To solve this problem, in this paper, we propose to \textbf{C}apture \textbf{A}ll-round \textbf{I}nformation \textbf{B}eyond \textbf{C}olor (\textbf{CAIBC}) via a jointly optimized multi-branch architecture for text-based person retrieval. CAIBC contains three branches including an RGB branch, a grayscale (GRS) branch and a color (CLR) branch. Besides, with the aim of making full use of all-round information in a balanced and effective way, a mutual learning mechanism is further employed to enable the three branches which attend to varied aspects of information to communicate with and learn from each other. Extensive experimental analysis is carried out to evaluate our proposed CAIBC method on CUHK-PEDES and RSTPReid in both \textbf{supervised} and \textbf{weakly supervised} text-based person retrieval settings, which demonstrates that CAIBC significantly outperforms existing methods and achieves the state-of-the-art performance on all the three tasks.
%
\begin{acks}
This work is partially supported by the National Natural Science Foundation of China (Grant No. 62101245, 61972016), China Postdoctoral Science Foundation (Grant No.2019M661999) , Natural Science Research of Jiangsu Higher Education Institutions of China (19KJB520009) and  Future Network Scientific Research Fund Project (Grant No. FNSRFP-2021-YB-21).
\end{acks}

\newpage

\bibliographystyle{ACM-Reference-Format}
\balance
\bibliography{jonniewayy}



\end{document}